# Vertical GaN Diode BV Maximization through Rapid TCAD Simulation and ML-enabled Surrogate Model

Albert Lu[1], Jordan Marshall[2], Yifan Wang[4], Ming Xiao[4], Yuhao Zhang[4], and Hiu Yung Wong[3*]
[1]Compter Engineering, [2]Aerospace Engineering, [3]Electrical Engineering, San Jose State University, CA, USA
[4]Center for Power Electronics Systems, Virginia Polytechnic Institute and State University, VA, U.S.A
*hiuyung.wong@sjsu.edu

**Abstract**

In this paper, two methodologies are used to speed up the maximization of the breakdown voltage (BV) of a vertical GaN diode that has a theoretical maximum BV of ~2100V. Firstly, we *demonstrated a 5X faster accurate* simulation method in Technology Computer-Aided-Design (TCAD). This allows us to find *50% more numbers of high BV (>1400V)* designs at a given simulation time. Secondly, a machine learning (ML) model is developed using TCAD-generated data and used as a surrogate model for differential evolution optimization. It can inversely design an *out-of-the-training-range* structure with BV as high as *1887V (89% of the ideal case)* compared to ~1100V designed with human domain expertise.

## 1. Introduction

GaN is becoming a mainstream semiconductor for RF and power applications, with a total market size of over $1 billion [1]. Vertical GaN devices, such as vertical GaN diodes, have been widely regarded as one of the most promising candidates for next-generation higher-voltage, high-power applications [2][3]. Due to its wide bandgap (3.4eV), the breakdown field of GaN is 10 times higher than that of Si. A GaN diode is expected to have more than 1450 times better Baliga's Figure-of-Merit than Si [2]. However, its full potential can only be unfurled if the diode has a proper edge termination, such as guard rings [4] and junction termination extension (JTE) [5], the design of which requires a lot of domain expertise and prolonged simulation time due to the huge design space. For GaN power diodes, the deployment of GaN-on-Si wafers can lower the cost but lead to higher leakage due to the threading dislocations [6][7]. GaN-on-GaN diode is preferred from the reliability considerations.

In this paper, two methods are proposed to speed up the simulation and to find high BV (>1400V) designs of GaN diode on GaN-substrate. One is to develop a faster but accurate TCAD simulation methodology. Another is to use TCAD-data-trained machine learning to enable surrogate model development to inversely design a high BV diode. Section 2 discusses the structure used. Section 3 discusses various approaches used for BV maximization and the results.

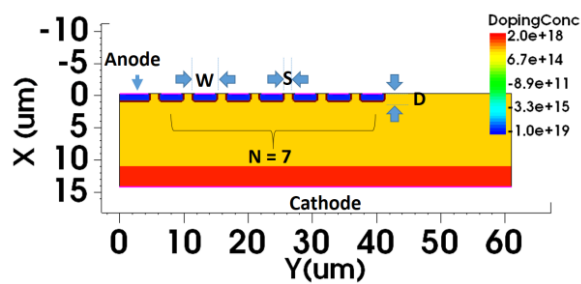

Figure 1: A simulation structure examplar of the vertical GaN diode used in this study. Guard ring number (N) = 7 is used as an example. Four of the five design variables (S, W, D, N) are highlighted. The standard deviation, σ, of the junction gradient is not shown.

## 2. Simulation Setup

TCAD Sentaurus is used for structure creation and device simulation [8]. Fig. 1 shows an example of the simulation structure. Guard rings are added next to the anode for edge termination. The drift region is 10μm with $10^{16}$cm$^{-3}$ n-type doping of silicon. The anode and the guard rings are p-type and doped with $10^{19}$cm$^{-3}$ magnesium. Fermi-Dirac statistics, incomplete ionization, high-field mobility saturation, and impact ionization are turned on using the calibrated parameter values in [9]. To maximize the BV, 5 design variables are used. The variables are the space (*S*) between the guard rings, the width (*W*) of the guard rings, the depth (*D*) of the guard rings, the number (*N*) of the guard rings, and the standard deviation (*σ*) of the guard ring junction. Fig. 2 shows the BV of an ideal 1D structure of about 2100V which represents the theoretical limit. The current is scaled by assuming the third dimension is 1mm. Using human expertise by experimenting with *S*, *W*, *D,* and *N*, the highest BV obtained is ~1100V.

## 3. Results and Discussion

*3.1. BV Maximization through TCAD Searching*

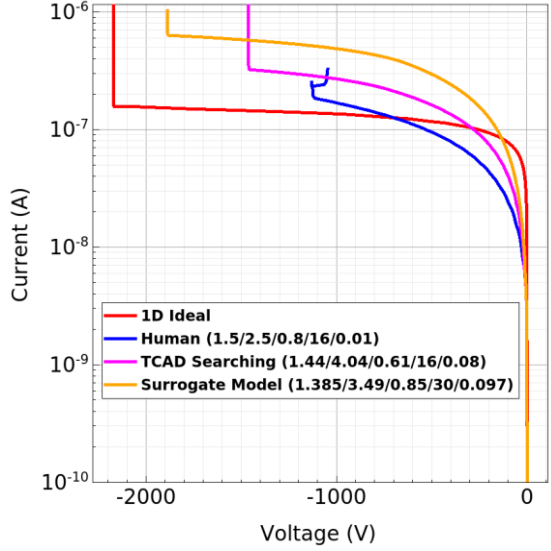

Figure 2: Reverse I-V curves of selected designs. The third dimension of the 2D structures is set to 1mm. The parentheses contain the values corresponding to S(μm), W(μm), D(μm), N and σ(μm).

We then generate various devices using TCAD by randomly creating structures with $S \in [0.25,5]$, $W \in [0.25,5]$, $D \in [0.01,1]$, $N \in [0,32]$, and $\sigma \in [0.01,0.1]$ all in μm except $N$ which is unitless. 300 structures were simulated, and the total simulation time is about 4 days on 30 cores. Only 2 structures are found to have BV > 1400V (the highest one is shown in Fig. 2) using this random TCAD searching method.

### 3.2. Rapid BV Simulation

It is desirable to find an *accurate* and fast simulation setup to speed up the TCAD searching process. Such a setup can also be used to generate enough data for machine learning (ML) in the following study. Various BV simulation simplification schemes such as removing incomplete ionization model, not solving hole continuity equation, using ionization integral method, etc. have been tested. Among them, it is found that only removing impact ionization and monitoring the peak electric field until it reaches 3.3MV/cm (GaN critical field) (fast model) provides a significant speedup and accurate solutions. Fig. 3 shows the relationship between the BV obtained using the fast model and the full model and they show a linear relationship.

Fig. 4 shows the distribution of simulation time of the two models. The speedup can be as much as 24 times and on average, the speedup is 5 times. Moreover, among the 300 simulations, *91% converge using the fast model and only 62% converge with the full model*.

The fast model is then used to simulate 3530 structures in about 5 days. In contrast, the full model could only simulate 300 structures in about 4 days. To compare the performance of the full and fast model, a comparison between the number of high BV (>1400V) structures obtained in each model is performed. If the fast model is more performant, then a larger amount of high BV structures would be expected. For the full model, it is already known that only 2 structures have high BV (>1400V). To check how many high BV structures are obtained in the fast model simulation, selected structures need to be verified using full model simulations. The following methodology is used to select the structures to be verified. Fig. 3 shows a 95% prediction interval used to determine this ideal search range of structures with the highest chance to become high BV (>1400V).

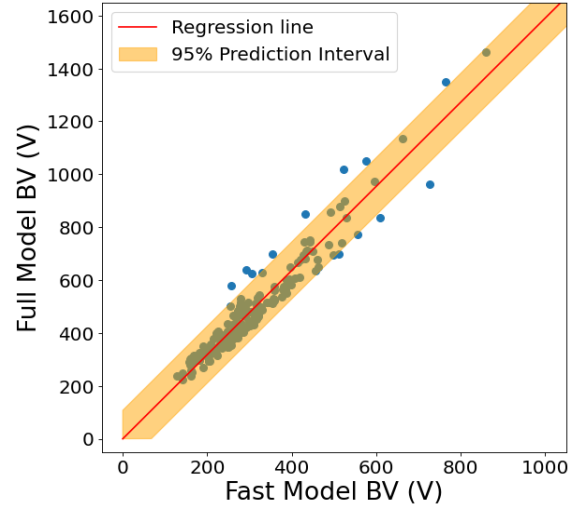

Figure 3: Relationship between the full model BV (with impact ionization) and the fast model BV (without impact ionization and extracting BV at maximum E-field = 3.3MeV/cm). The fitted slope is 1.5893.

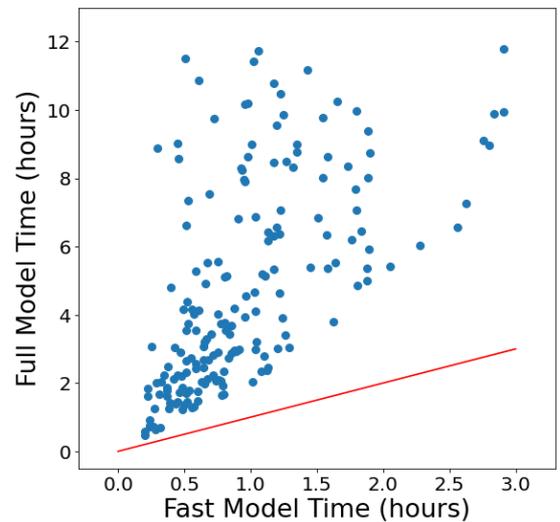

Figure 4: Comparison of the simulation time using the full and fast models. Each point represents one simulated structure. The red line is y = x.

This range is found to be above 812 V in the fast model BV. There are 7 structures in this range and full model simulations of these structures are performed for further verification. 3 structures out of the 7 are found to have high BV (>1400V). Thus, this result is 50% more than the amount found solely using the full model in a similar amount of time.

The benefit of the fast model can be seen by how it expands the search space greatly when the threshold of "high BV" is defined with lower values. Table I shows different redefinitions of what is considered "high BV" and the resulting performance. The amount of "high BV" in the full model still stagnates even if "high BV" is redefined with lower bounds. In contrast, the search space for the fast model increases greatly as "high BV" is redefined with lower bounds. This large search space is due to the speedup of the fast model, which allows for significantly more structures to be simulated at a reasonable time.

TABLE I

DIFFERENT DEFINITIONS OF "HIGH BV" AND NUMBER OF "HIGH BV" OBTAINED

| "High BV" | Full Model | Fast Model Search Space | Fast Model Actual "High BV" | Gain by using Fast Model |
|---|---|---|---|---|
| 1400 | 2 | 7 | 3 | 1.5X |
| 1300 | 3 | 12 | 5 | 1.67X |
| 1250 | 3 | 17 | 7 | 2.33X |

### 3.3. BV Maximization by ML-enabled Surrogate Model

Even though the fast model can speed up the simulation by 5X, it still cannot find a design with a high enough BV. With the speed-up gained by the fast model, lots of data are made available to train ML models quickly. Two ML models are thus developed to correlate the 5 design parameters to the BV of the two datasets generated with the fast model earlier. Thus, the goal of this model is to predict the BV of a structure as if it were run by a fast model simulation, not the full model. The models are called NN275 and NN3530, where one model is trained with 275 structures and the other is trained with 3530 structures, respectively. Keras [8] was used to train the models. Each model is a neural network (NN) with 1 input layer, 2 hidden layers (each has 50 hidden nodes with L2 regularization followed by batch normalization), and 1 output layer is used (Fig. 5). 80% of the data is used with cross-validation and 20% of the data is used for testing. 10-fold cross-validation is used with 3 repeats for hyperparameter tuning and training with scikit-learn [9]. With the final test set, NN275 obtains $R^2$ above 0.77 and NN3530 obtains $R^2$ above 0.95.

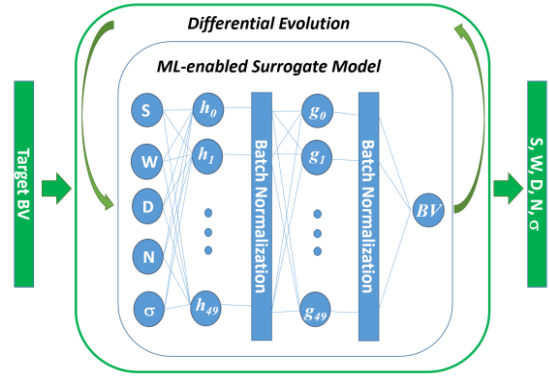

Figure 5: NN used as a surrogate model for differential evolution algorithm to design guard rings for a target BV.

The trained machines are then used as surrogate models for the differential evolution algorithm to design the guard ring based on any target BV. This is achieved by minimizing $|f(S, W, D, N, \sigma) - V_{target}|$, where $f$ is the output of the ML surrogate model and $V_{target}$ is the target BV when running the fast model. $V_{target}$ from 0V to 1250V are then fed into the differential evolution algorithm to inversely design the diode for the given $V_{target}$. Each of the optimizations takes only ~30 minutes on a laptop. The Python library, SciPy [10], was used to run the differential evolution algorithm. The algorithm is a population-based method and does not use gradients to minimize.

For example, for $V_{target}$ = 1050V, it is deduced that $S = 1.385, W = 3.49, D = 0.85, N = 30, \sigma = 0.097$ should be used. The corresponding TCAD structures are then constructed and simulated with the full model. Fig. 6 shows that the algorithm can inversely design the device well and most of the BV is near or within the expected range. The expected range is from the 95% prediction interval in Fig. 3 which shows the variance in scaling between the fast and full models. Some points likely lie outside the expected range due to the ML model being trained on the fast model datasets, which are expected to have some inaccuracy in predicting a structure with the given $V_{target}$ and with scaling to the full model.

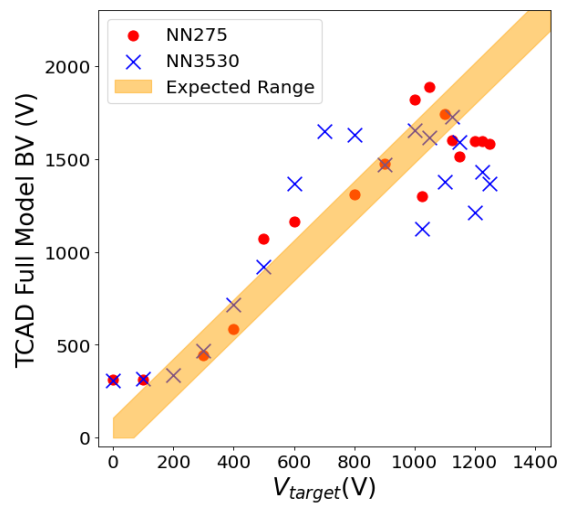

Figure 6: TCAD simulatated BV for the inverse designed structures predicted by the differential evolution trained on NN275 and NN3530. Expected range is based on the relationship found in Fig. 3.

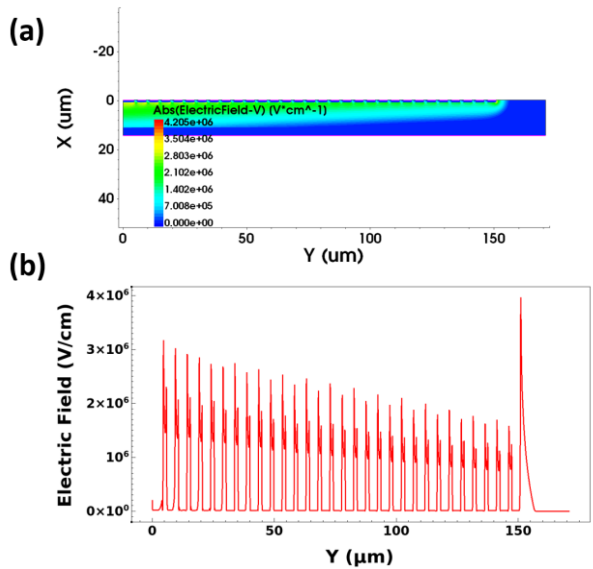

Figure 7: Electric field distribution of the structure that achieves the highest BV (1887 V). Top: 2D plot. Bottom: 1D cut line along the junction bottom.

Fig. 6 also shows how NN3530 has 7 structures within the expected range and thus is found to be slightly more accurate than NN275. However, NN275 still has many of its structures close to the expected region and thus both methods are found to be closely matched. This implies that NN275, which was trained on only 275 structures, has sufficient data for optimization. It can also achieve BV = 1887V for $V_{target}$ = 1050V (target of the fast model), much higher than its training data after scaling (Fig. 2). This is ~89% of the ideal value. Fig. 7 shows the electric field distribution of this structure. NN275 is believed to be as performant as NN3530 even with less data it is trained on because it seems to be able to capture the underlying patterns sufficiently.

## 4. Conclusion

We proposed a new TCAD setup that has a 5X faster speed and 2X better convergence for GaN diode BV simulation and has a linear correlation to the full model BV. This allows the finding of 50% more designs with high BV (>1400V). To further explore the design with higher BV, two NNs are built as surrogate models and, by using differential evolution, a design with BV as high as 1887 V is discovered. Both NNs are comparable, and it is found that both are similar in performance.

## Acknowledgment

This material is based upon work supported by the National Science Foundation under Grant No. ECCS-2134374.